\documentclass[preprint,12pt]{elsarticle}

\usepackage{amssymb}
\usepackage{amsthm}
\usepackage{amsfonts}
\usepackage{amsmath}

\usepackage[normalem]{ulem}

\usepackage{multirow}
\usepackage{xcolor}
\usepackage{float}
\usepackage{url}

\journal{Neurocomputing (\url{10.1016/j.neucom.2023.01.059})}

\begin{document}

\begin{frontmatter}



\title{Density-based clustering with fully-convolutional networks for crowd flow detection from drones
}


\author[inst1]{Giovanna Castellano}
\author[inst1]{Eugenio Cotardo}
\author[inst1]{Corrado Mencar}
\author[inst1]{Gennaro Vessio}

\affiliation[inst1]{organization={Department of Computer Science, University of Bari Aldo Moro},
            city={Bari},
            country={Italy}}



\begin{abstract}
Crowd analysis from drones has attracted increasing attention in recent times due to the ease of use and affordable cost of these devices. However, how this technology can provide a solution to crowd flow detection is still an unexplored research question. To this end, we propose a crowd flow detection method for video sequences shot by a drone. The method is based on a fully-convolutional network that learns to perform crowd clustering in order to detect the centroids of crowd-dense areas and track their movement in consecutive frames. The proposed method proved effective and efficient when tested on the Crowd Counting datasets of the VisDrone challenge, characterized by video sequences rather than still images. The encouraging results show that the proposed method could open up new ways of analyzing high-level crowd behavior from drones.
\end{abstract}




\end{frontmatter}


\section{Introduction}
Due to population growth and the increasing degree of urbanization, more and more people live in urban areas. Positive consequences of this trend are the enrichment of cultural life and the full use of convenient urban infrastructure. At the same time, gatherings of people, which can occur for various reasons, such as political demonstrations, festival celebrations, concerts, and so on, pose serious challenges to urban security and management. In this perspective, automated crowd analysis methods, which typically involve crowd counting and associated crowd density estimation, have attracted increasing attention for their many potential applications~\cite{sindagi2018survey,li2021approaches}. These include the prevention of crowd-induced disasters, such as stampedes, but also other less critical objectives such as better crowd management at public events and the design of public spaces and virtual environments.

A cost-effective way to perform automated crowd analysis is by using unmanned aerial vehicles (UAVs), more commonly known as drones. Indeed, once equipped with affordable but sufficiently powerful cameras and GPUs, drones can become flying computer vision devices that can be rapidly deployed for a wide range of applications, including crowd analysis for public safety~\cite{akbari2021applications}. However, while these perspectives are fascinating, there are also some drawbacks to be aware of. On the one hand, the computer vision algorithms applied to aerial images are burdened with further difficulties because the problems of scale and point of view are taken to the extreme. On the other hand, the sophisticated and computationally intensive methods commonly applied in this field do not meet the stringent real-time requirements imposed by the UAV. In other words, lightweight models that offer a good compromise between effectiveness and efficiency are essential~\cite{tzelepi2019graph}.

Crowd analysis with drones has attracted attention in recent years~\cite{zhu2021detection}. However, despite significant progress, the proposed methods still have room for improvement to address the challenges posed by drones. 
In this article, we want to contribute to this research effort by taking it one step further: instead of considering crowd counting and density estimation in static frames, we aim to detect crowd flow. This poses a new challenge as the goal is not only to recognize the presence of people in a single high-altitude scene but also to determine how crowds flow as a function of time. This is different from people tracking---where the goal is to track a single person or groups of people---and can lead to useful systems, as it can allow for crowd behavior analysis for better logistics and disaster prevention~\cite{kok2016crowd}.

To this end, we propose a method for crowd flow detection from drones based on fully-convolutional networks (FCNs). The network is trained to recognize groups of people in each frame and, to do this, simultaneously learns to perform crowd density estimation and crowd clustering. In this way, groups of people are identified simply by their centroids, and these are used to trace the trajectories of the identified groups, following their movement during the shooting of the drone. We preferred FCNs over other architectures mainly because of their known efficiency. Furthermore, direct learning of crowd clusters was preferred to avoid a multi-step approach, based on performing density estimation first and clustering the resulting density maps later. This was done in our previous preliminary work~\cite{IJCNN2022} but, while effective, this approach proved too demanding from a computational point of view. The method was tested on the recently proposed Crowd Counting 2020~\cite{du2020visdrone} and 2021~\cite{liu2021visdrone} datasets, used annually for the international VisDrone challenge. The peculiarity of these datasets is that they are not characterized by still images but actually by frames of video sequences that are used here to perform the crowd flow detection task. \textcolor{black}{It is worth noting that what we actually do to determine crowd flow is to calculate the inter-frame difference between centroids; in other words, it is a kind of ``inter-frame density clustering''. However, since this approach allows us to detect the movement of the crowd frame by frame, we use the expression ``flow detection'' for the sake of simplicity.}

The rest of this paper is structured as follows. Section~\ref{sec:related} reviews related work. Section~\ref{sec:materials} describes the datasets used. Section~\ref{sec:methods} presents the proposed method. Section~\ref{sec:experiments} describes the experimental setup and discusses the results obtained. Section~\ref{sec:conclusion} concludes the paper and highlights the future developments of our research.

\section{Related work}
\label{sec:related}
There is a large body of knowledge about crowd counting and crowd density estimation in computer vision, but the trend today is density estimation. Early work usually applied a person or head detector via a sliding window on the image. However, although current implementations may be based on state-of-the-art object detectors such as YOLO~\cite{lan2018pedestrian,molchanov2017pedestrian}, these approaches still provide unsatisfactory results when asked to detect small objects in a very dense crowd. To alleviate this problem, regression-based methods have been introduced that directly learn the mapping from an image to the global people count~\cite{sindagi2018survey}. However, although these methods make the approach independent of the precise position of individuals in the crowd, which is very complex, they ignore the spatial information that can be very useful for prediction. To avoid the difficulty of accurately detecting and locating people in the scene, while using spatial information, the recent trend is to learn density maps, thus incorporating spatial information directly into the learning process~\cite{gao2020cnn}. Successful solutions include methods that work first at the patch level and then fuse local features~\cite{zhu2020crowd}, methods that integrate attention mechanisms~\cite{zhang2020cross}, cascade approaches that jointly learn people counting and density maps~\cite{sindagi2017cnn}, \textcolor{black}{methods that improve performance through knowledge distillation~\cite{jiang2021shufflecount}, and frameworks that simultaneously perform crowd counting and localization~\cite{jiang2021smartly}.}

However, while effective, these approaches are generally computationally demanding and do not meet the stringent requirements typically imposed by UAVs (limited battery, need for real-time responses). How to fine-tune deep neural architectures to achieve an optimal balance between precision and performance is an active research area. The VisDrone Crowd Counting challenge was introduced to encourage research in this direction~\cite{du2020visdrone,liu2021visdrone}; nevertheless, the solutions proposed by the participants in the challenge are not always focused on efficiency but rather on effectiveness, as the goal is only to obtain a low error in counting people. The lowest error was obtained with TransCrowd~\cite{liang2021transcrowd}, based on the increasingly popular Vision Transformer~\cite{dosovitskiy2020image}. However, the proposed method only regresses the people count, not providing density maps that would be useful for detecting crowd flow; furthermore, transformer-based solutions are known to be computationally expensive.

A promising way to address these problems is to use FCN models. Because they do not rely on fully-connected layers at all, which are the most expensive part of processing a neural network, they are a candidate solution for finding an accurate model without damaging the inference time. An FCN model for aerial drone imaging was presented in~\cite{tzelepi2019graph}, and a similar solution was also proposed in our previous work~\cite{castellano2020crowd}. However, both methods were aimed at crowd detection, i.e.~discriminating between crowded and uncrowded scenes; furthermore, they only provide coarse density maps, as the models have not been trained on people labels.

Human tracking methods based on RGB cameras or other sensors that use clustering or classification models to track motion have been investigated, e.g.~\cite{gajjar2017human,xiao2019human,yan2020online}. However, they are designed to work indoors or by involving a few people from a frontal perspective. The work most linked to ours, which takes into account the images captured by drones, is~\cite{wen2021detection} where the same authors who propose the VisDrone Crowd Counting datasets present a model that jointly solves density map estimation, localization, and tracking. This model differs from ours in that it uses a complex and expensive pipeline aimed at tracking individual trajectories. Other authors have recently proposed a method for periodic crowd tracking from UAVs based on a binary linear programming model~\cite{chebil2022toward}. However, they experimented with simulated scenarios that do not consider the crowd detection problem from a computer vision perspective. As far as we know, there is no work in the literature addressing crowd flow detection in drone videos, which poses significantly different challenges than traditional settings. This paper aims to fill this gap; in particular, we aim to trace the centroids that identify groups of people by exploiting the spatial information learned and expressed through density maps.

\section{Materials}
\label{sec:materials}
The literature landscape is not as populated with datasets for crowd counting and density estimation from drones with video sequences captured by optical cameras. The datasets best suited to our purposes for evaluating crowd flow detection were the VisDrone Crowd Counting 2020~\cite{du2020visdrone} and 2021~\cite{liu2021visdrone} datasets. The two benchmarks are characterized as follows:
\begin{itemize}
    \item VisDrone Crowd Counting 2020 (CC2020) contains 82 \textcolor{black}{video clips} ($2,460$ frames in total) with a resolution of $1920 \times 1080$;
    \item VisDrone Crowd Counting 2021 (CC2021) contains $1,807$ frames with a resolution of $640 \times 512$. Unlike CC2020, \textcolor{black}{the frames are not arranged in precisely separated video clips}, and this required manual separation to split the sequences of different locations to avoid overfitting.
\end{itemize}

The video sequences were acquired by various drone-mounted cameras, which shot different scenarios in different cities in China to maintain diversity. In both datasets, people were annotated manually with dots in each video frame, expressed as $(x, y)$ coordinates in the bi-dimensional plane. However, CC2020 and CC2021 have significant differences. In CC2020, each of the 82 sequences was captured by a drone hovering over the crowd, allowing for rather static scenes. In CC2021, on the other hand, the drone flies and sometimes rotates, shooting different scenes even if semantically linked. Furthermore, it should be noted that in CC2020 there is a predominance of daylight scenes, while in CC2021 there are many frames at night, making the dataset much more variable in this respect. A final difference concerns altitude: in CC2020 the frames appear to have been shot at a higher altitude than in CC2021. For both datasets, we randomly held out (sequence-wise) a fraction of 25\% of the total frames evenly split to form validation and test sets.

Compared to benchmark datasets focused on crowd counting in surveillance scenes, both VisDrone datasets present particular challenges due to the scenes captured by drones. Object scales can be extremely small due to the high shooting altitude of drones. Crowds are sparse across the video frames, as each can hold a few to dozen people. Finally, the crowds are surrounded by very different backgrounds in different sequences.

\section{Methods}
\label{sec:methods}
The proposed method for crowd flow detection is based on an FCN model, which is used to estimate a ``centroid density map''---a heatmap highlighting crowd centroids---\textcolor{black}{from pairs of consecutive frames of the same video sequence} shot by the drone. Since a crowd can be seen as multiple groups of people not necessarily following the same direction, the predicted centroids are representative of these groups of people. The displacements of the centroids detected \textcolor{black}{between the pairs of frames} are then calculated to identify the direction of movement. \textcolor{black}{Thus, the framework assumes video sequences shot by a drone, but the network is fed one frame at a time.}

\textcolor{black}{The idea of combining a density estimation method with clustering, instead of tracking the movement of each individual, is motivated by the complexity and computational cost of this strategy. In fact, while the direct use of the crowd density map based on the location of all individuals can retain more information, it also carries the burden of tracking the trajectory of each individual in the scene. Moreover, individual tracking can be not only impractical but also non-essential, as in crowd management scenarios it is important to recognize the overall flow of people rather than the precise location of each person in the scene. A centroid density estimation method will focus only on high-density areas, i.e.~those corresponding to concentrations of people, and will be inherently robust to occlusion, which would heavily affect a people detector, especially from a high altitude.}

\begin{figure}[t]
    \centering
    \includegraphics[width=\textwidth]{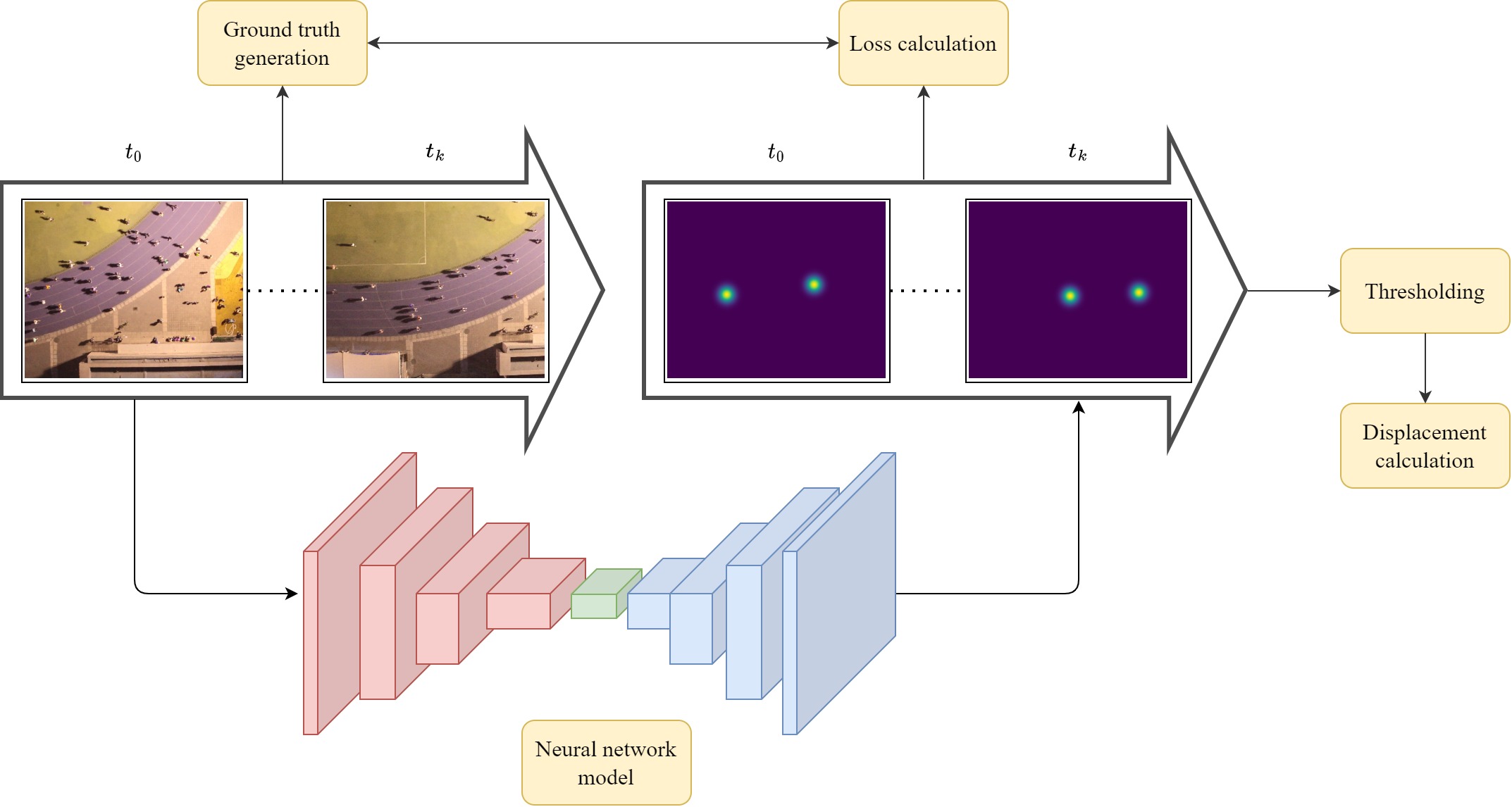}
    \caption{\textcolor{black}{Schema of the overall method. Pairs of consecutive frames taken by a drone at time $t_0$ and $t_k$ (with $t_k > t_0$ and separated by any time interval) are fed into the neural network model for crowd clustering. A synthetically generated ground truth helps guide the learning of the network. The obtained centroid density maps are then thresholded, and the displacement of the centroids, representing the groups of people, from $t_0$ to $t_k$, is used to determine their direction of movement.}}
    \label{fig:process}
\end{figure}

The method is detailed in the following subsections, along with an explanation of how the ground truth is obtained. \textcolor{black}{An illustration schematizing the overall processing is shown in Fig.~\ref{fig:process}.}

\subsection{Ground truth generation}
\begin{figure}[t]
    \centering
    \includegraphics[width=.3\textwidth]{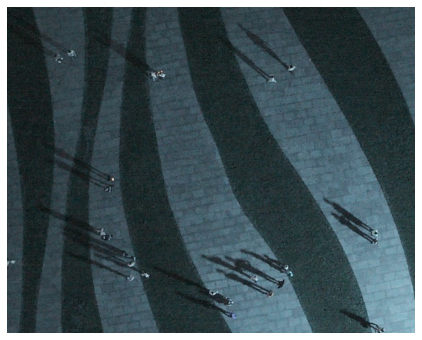}
    \includegraphics[width=.3\textwidth]{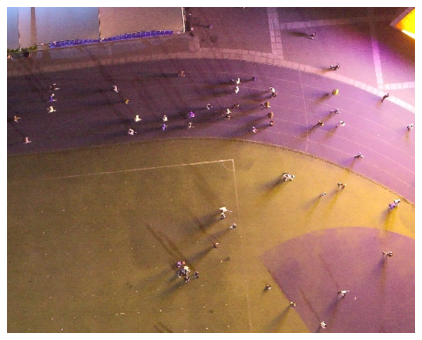}
    \includegraphics[width=.3\textwidth]{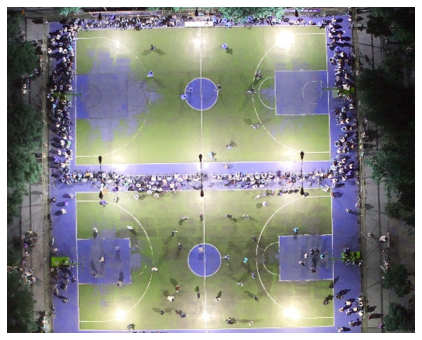}
    \includegraphics[width=.3\textwidth]{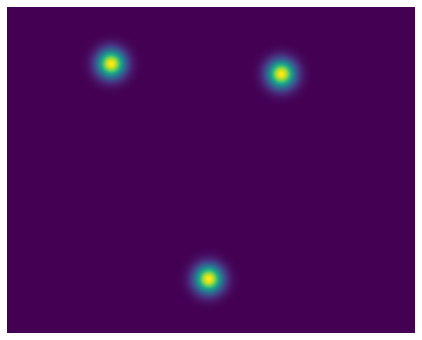}
    \includegraphics[width=.3\textwidth]{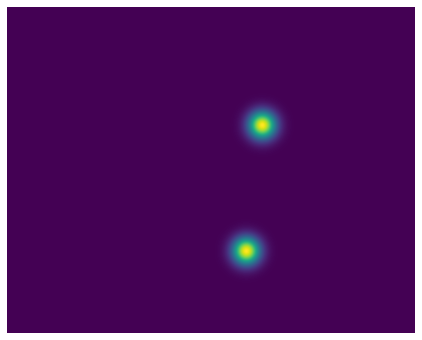}
    \includegraphics[width=.3\textwidth]{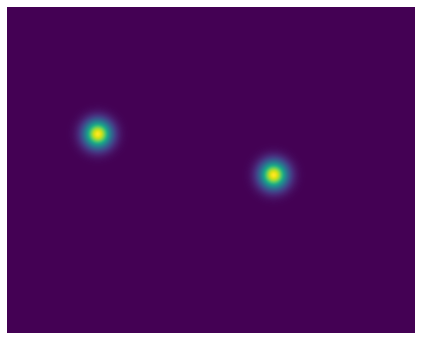}
    \caption{\textcolor{black}{Examples of frames and corresponding synthetically generated centroid density maps.}}
    \label{fig:ground_truth}
\end{figure}

Since we assume that there is no label regarding the location of the centroids within the original drone shot, we followed a simple strategy to derive a ground truth for the crowd centroids. We first applied the well-known Mean Shift clustering algorithm to the head annotations described above to obtain centroids. Mean Shift is a centroid-based algorithm that updates the candidates for the centroids as the average of the points within a given region. These candidates are then filtered in a post-processing step to eliminate near-duplicates~\cite{comaniciu2002mean}. \textcolor{black}{Specifically, the Scikit-learn implementation of the Mean Shift algorithm with the recommended default hyperparameters was used.\footnote{At the time of writing, the version of Scikit-learn is 1.1.2.}} Then, following the seminal paper by Zhang et al.~\cite{zhang2016single}, if there is a centroid at pixel $x_i$, we represented it as a delta function $\delta(x - x_i)$ and we obtained a ``centroid density map'' $C(x)$ convolving the delta function with a Gaussian kernel:
\[
C(x) = \sum_{i=1}^{K} \delta (x - x_i) * G_{\sigma}(x)
\]
In the formula, $K$ is the number of centroids; $\delta$ equals $1$ when $x = x_i$, $0$ otherwise; and $G_{\sigma}$ is the Gaussian kernel. Since the sizes of the heads are similar in each video sequence and there is no perspective problem as in \cite{zhang2016single}, we decided to use a fixed $\sigma$ for each frame. In particular, we have empirically set $\sigma = 10$, since this value leads to better performance thanks to the sufficiently large ``confidence'' activation area. \textcolor{black}{Examples of generated centroid density maps are shown in Fig.~\ref{fig:ground_truth}.}

We used Mean Shift as it is impossible to know the number of clusters in the crowd in advance, so there was a need for an algorithm that did not require a pre-specification of the number of clusters. In this way, an unsupervised learning approach is followed to find crowd centroids, but when their location is found, it can be used as a synthetically generated annotation to guide a supervised learning approach. Although this strategy strongly depends on the results provided by the specific clustering algorithm chosen, it allowed us to automate the generation of the ground truth and to guide and quantitatively evaluate the clustering task performed by the neural network.

In addition, we also experimented with models aimed at performing the more classic crowd density estimation task. To this end, ``crowd density maps'' $D(x)$ were obtained by convolving the same delta function with Gaussian kernel as before but using the original people head annotations.

\subsection{Fully-Convolutional Networks}
The proposed method is based on an FCN architecture that recognizes the crowd centroids within each frame and produces the related heatmap. In our previous preliminary work~\cite{IJCNN2022}, the identification of the centroids was delegated to a classic clustering algorithm \textit{after} the generation of crowd density maps, significantly penalizing the inference time. In the method proposed here, instead, the task of finding crowd centroids is integrated directly into the network training and is performed in a single step. 

Fully-convolutional neural networks, originally proposed in~\cite{long2015fully}, perform only convolution and pooling operations and discard the fully-connected component typical of CNNs. Instead of the fully-connected layer, there is a $1 \times 1$ convolution with stride $1$, which allows, on the one hand, to have fewer parameters to estimate and, on the other hand, to be able to receive an image of arbitrary size as an input. It is worth noting that although FCNs have this desirable property, for better evaluation, we have set the input resolution of each frame to $640 \times 512 \times 3$. The architecture of an FCN consists of two parts: an encoder aims to downsample the input into a lower-dimensional representation, a decoder aims to upsample the latent representation to the desired output resolution. 

In particular, \textcolor{black}{we have experimented with two different neural networks: an \textit{ad-hoc} FCN designed to be as simple as possible, and a state-of-the-art FCN already used for crowd counting in more traditional contexts, i.e.~MobileCount~\cite{wang2020mobilecount}.} In the ad-hoc implementation, a rescaling layer normalizes each pixel value in the $\left[ 0, 1 \right]$ range. Then, the encoder part of the model repeatedly applies four blocks consisting of a convolutional layer with kernel $3 \times 3$, batch normalization, and max pooling with kernel $2 \times 2$ until reaching a latent space $\mathsf{Z} \in \mathbb{R}^{40 \times 32 \times 128}$. The decoder upsamples the feature maps with transposed convolution and batch normalization. Finally, a $1 \times 1$ convolutional layer produces the output density map of size $640 \times 512 \times 1$. The commonly used ReLU was chosen as the activation function. \textcolor{black}{MobileCount, on the other hand, is similar but has some more sophisticated architectural designs. To reduce the input resolution, a $3 \times 3$ max pooling layer with stride 2 is added before the encoding part. The encoder is adapted from MobileNetV2~\cite{sandler2018mobilenetv2} by reducing the number of inverted residual blocks from 7 to 4. As for the decoding component, the lightweight RefineNet~\cite{nekrasov2018light}, originally designed for semantic segmentation, is exploited.}

\begin{figure}[H]
    \centering
    \includegraphics[width=.85\textwidth]{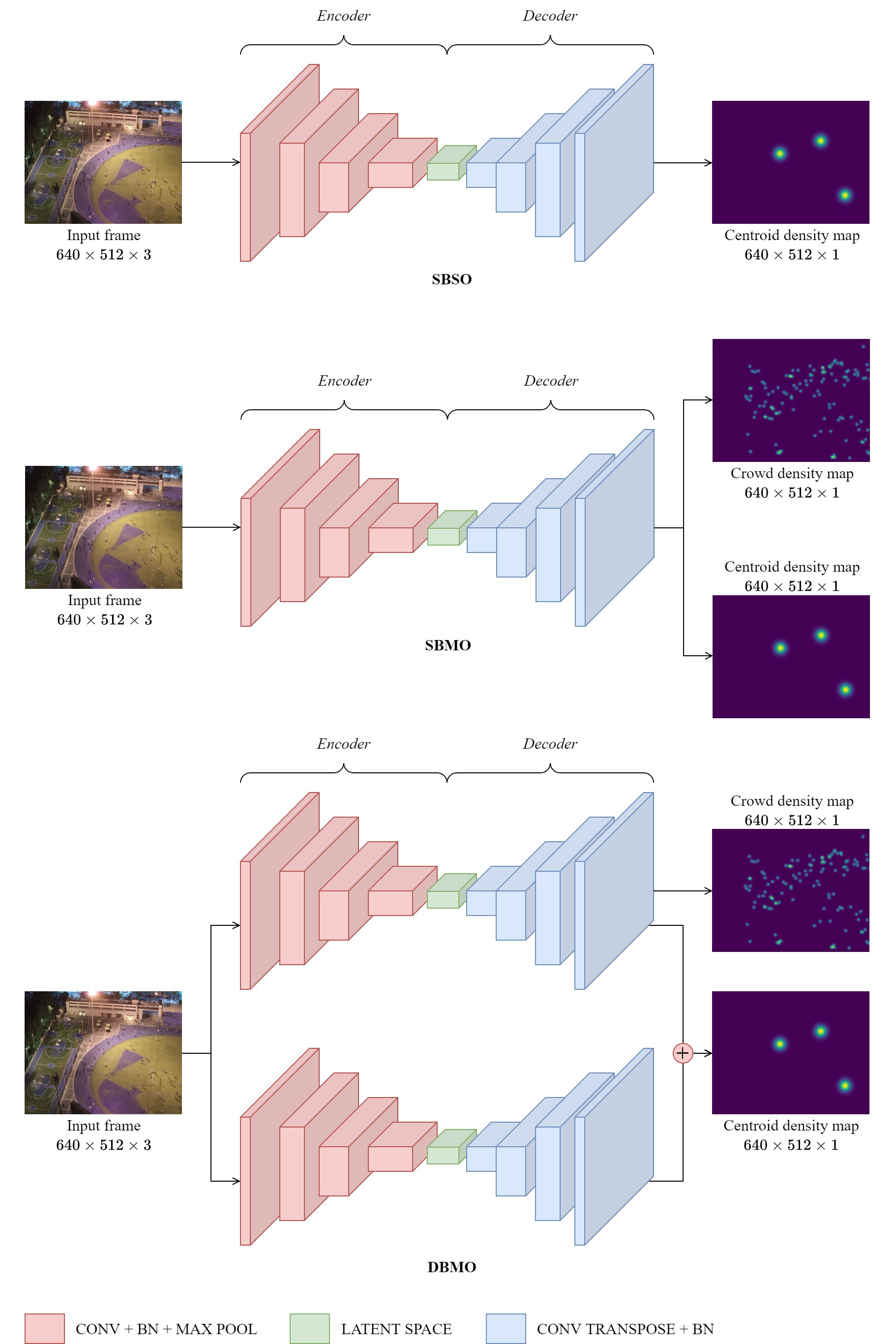}
    \caption{\textcolor{black}{Schema of the proposed neural network architectures. The proposed ad-hoc FCN is schematized in this figure but can be replaced by any similar model as desired.}}\label{fig:models}
\end{figure}

Since we have two datasets with very different characteristics, as shown in Fig.~\ref{fig:models} we experimented with three main variants of the above architectures, which were likely to produce different results:
\begin{itemize}
    \item \textit{Single-branch, single-output} (SBSO): \textcolor{black}{this architecture uses one of the two FCNs described above} and learns to directly estimate the centroid density map. As a loss function, the network is trained to minimize the mean squared error between the predicted and the ground truth centroid density map:
    \[
    \mathcal{L} = \frac{1}{N} \sum_{i=1}^N \left \| C^P(i) - C^{GT}(i) \right \|_2^2
    \]
    where $N$ is the number of samples, $C^P(i)$ and $C^{GT}(i)$ are the predicted and ground truth centroid density map, respectively, and $\left \| \cdot \right \|_2$ is the Euclidean distance. Demanding the model to find groups of people in a completely unsupervised way would not have provided the network with a guide to optimize weights to cluster only a specific type of objects, in our case people, leaving it completely free to separate people from trees, buildings, etc., which was not our goal.
    \item \textit{Single-branch, multi-output} (SBMO): The task of finding crowd centroids may be difficult for the model as the centroids do not represent objects effectively present within the scene. To try to mitigate this issue, we also experimented with a multi-output model, which extends the previous one by learning to simultaneously estimate the centroid density map and the more classic crowd density map. The main idea is that the auxiliary task can help the overall network extract the features that actually characterize the people in the scenes, thus supporting the main task of density estimation, which only concerns the detection of crowd centroids. In this case, the loss function is the sum of the above and the mean squared error between the predicted and the ground truth crowd density maps.
    \item \textit{Dual-branch, multi-output} (DBMO): finally, to improve the capacity of the previous variant, we also experimented with a multi-output model, this time characterized by two branches, which are actually two ``twin'' autoencoders with the same architecture as before (\textcolor{black}{ad-hoc FCN or MobileCount}). The two branches communicate with each other through a concatenation of the feature maps produced immediately before the desired output: one branch is totally dedicated to learning the crowd density map; the concatenation of both branches contributes to estimating the centroid density map. The loss function is the same as the previous variant.
\end{itemize}
In other words, the multi-output architectures are meant to help the model learn more about the concept of ``crowdedness''. 
\textcolor{black}{Because these variants inherently perform crowd counting, in addition to crowd clustering, it is worth noting that they can be used for the more classical task of pedestrian counting if needed.}

\textcolor{black}{Finally, it is important to emphasize that the neural network models we have tested are a backbone of the overall framework we propose, which can be replaced with similar models as desired. Table~\ref{tab:parameters} compares the two backbones used in this work in terms of the number of parameters.} \textcolor{black}{The proposed FCN has been designed as a classic and straightforward baseline. It is small, but deliberately without any sophistication to further improve effectiveness or efficiency, to show the feasibility of the proposed method with a simple model. MobileCount, on the other hand, has more parameters but was purposely made by the authors as an efficient framework for real-time crowd counting. In fact, the MobileNetV2-based encoder and the RefineNet-based decoder were carefully adapted to achieve a good balance between accuracy and speed, as reported in the article~\cite{wang2020mobilecount}.}

\begin{table}[t]
\centering
\scalebox{0.7}{
\begin{tabular}{l|c|c|}
\cline{2-3}
                           & Ad-hoc FCN & MobileCount \\ \hline
\multicolumn{1}{|l|}{SBSO} & 0.35           & 25.199            \\ \hline
\multicolumn{1}{|l|}{SBMO} & 0.35           & 25.199            \\ \hline
\multicolumn{1}{|l|}{DBMO} & 0.7           & 50.398            \\ \hline
\end{tabular}}
\caption{\textcolor{black}{Comparison in terms of the number of parameters, in millions, between the proposed ad-hoc FCN and MobileCount, used as the backbone of the proposed architectures.}}\label{tab:parameters}
\end{table}

\subsection{Crowd flow detection}
Although accurate, FCN-predicted density maps are characterized by non-normalized values and may contain noise. In particular, while in the ground truth the background has a value of 0 for construction, there is no guarantee that its value will remain 0 in the predicted heatmap. To study how this affects performance, we first apply a min-max normalization to limit their range to $\left[0, 1 \right]$; then, we threshold (with an empirically chosen threshold $\tau$) the pixel values, so that any value below the threshold is considered to be the background. A higher threshold essentially maintains the areas where the model is more confident about the presence of people in those areas.

After identifying the centroids and filtering the images through the empirical threshold $\tau$, to calculate the actual displacement of the recognized groups of people, and to determine the direction of their movements, we calculate the difference in coordinates of the centroids predicted in different frames. In other words, the displacement of the centroids in an ending frame at time $t_k$ is calculated with respect to the centroids detected in a starting frame at time $t_0$, and so on. Since we are in two dimensions, this shift is simply calculated using the Euclidean distance between the $(x, y)$ coordinates. The primary assumption is that a given group of people, i.e.~a centroid in this case, can potentially move between successive frames, but its distance from its position in the previous frame would be less than the distance from all other centroids in the current frame. Therefore, to determine if we can associate the newly predicted centroids to the existing ones, and thus in which direction they have moved, we calculate the Euclidean distance between each pair of centroids in each frame and keep the minimum distances between the pairs to match them. In this way, we can classify each shift as facing one of the four typical cardinal points, namely North, South, East, and West, plus the intermediate points North-East, North-West, South-East, and South-West. The case in which the centroids remain stationary between the two frames is also considered. 

If $P_i$ is the set of centroids predicted at time $i$, we can distinguish three scenarios:
\begin{itemize}
    \item $\left| P_0 \right| = \left| P_k \right|$: in this case, each centroid in $P_0$ is simply associated with the \textit{closest} centroid in $P_k$;
    \item $\left| P_0 \right| > \left| P_k \right|$: in this case, we have fewer centroids in the ending frame. Indicated by $d = \left| \left| P_0 \right| - \left| P_k \right| \right|$ the difference in absolute value between the number of centroids in $t_0$ and in $t_k$, we will have that $d$ centroids in $t_0$ will remain unmatched. This could indicate a network prediction error, the fusion of two centroids in $t_0$ into one in $t_k$, or centroids leaving the visual field at time $t_k$;
    \item $\left| P_0 \right| < \left| P_k \right|$: in this case, we have fewer centroids in the starting frame. We will then have $d$ more centroids at time $t_k$, which will remain unmatched and could be indicative of a network prediction error, the formation of new clusters in $t_k$, or finally the input into the visual field of new centroids.
\end{itemize}

\section{Experiments}
\label{sec:experiments}
The proposed method was implemented in Python, using TensorFlow for the implementation of the deep learning models.\footnote{\url{https://github.com/evgenivs/crowd_flow_detection_drones}} All models were trained with stochastic gradient descent with the Adam optimizer and a learning rate of $10^{-4}$. As for the $\tau$ threshold, we experimented with values ranging from 0 to 1, reporting those that gave the best performance. As mentioned above, all frames have been resized to $640 \times 512$. All experiments were performed on Google Colab Pro, which mainly provides a T4 or P100 GPU. The performance metrics considered, as well as the results obtained, are described below.

\subsection{Performance metrics}
Applying traditional \textit{external} clustering measures is not feasible in our density-based context as they assume an exact match between discretized category labels. Since we are interested in measuring how correctly the detected centroids represent relevant groups of people within the scene, we propose a new ad-hoc metric (already presented in~\cite{IJCNN2022}) which we called \textit{Mean Coordinate Matching Error} (MCME). The metric measures the average distances between the ground truth centroids and the predicted centroids. For a single frame, let:
\begin{equation*}
A = 
\left\{
  \left(C^{P},C_{\min}^{GT}\right) \;\middle|\;
  \begin{aligned}
  & C^{P}\in P,\\
  & C_{\min}^{GT}=\arg\min_{C^{GT}\in GT}\left\Vert C^{P}-C^{GT}\right\Vert _{2}
  \end{aligned}
\right\}
\end{equation*}
\begin{equation*}
B = 
\left\{
  \left(C_{\min}^{P},C^{GT}\right) \;\middle|\;
  \begin{aligned}
  & C^{GT}\in GT,\\
  & C_{\min}^{P}=\arg\min_{C^{P}\in P}\left\Vert C^{P}-C^{GT}\right\Vert _{2}
  \end{aligned}
\right\}
\end{equation*}
where $GT = \{ C_1^{GT}, C_2^{GT}, \ldots \}$ and $P = \{ C_1^P, C_2^P, \ldots \}$ are the ground truth and the predicted centroids, respectively. Then:
\[
MCME=\frac{1}{\left|A\cup B\right|}\sum_{\left(C^{P},C^{GT}\right)\in A\cup B}\left\Vert C^{P}-C^{GT}\right\Vert _{2}
\]
Each centroid of a set (say $P$) is associated with the nearest neighbor centroid of the other set ($GT$) since both sets of centroids are assumed to represent the same structure in the data. This association must be symmetric, i.e.~from $P$ to $GT$ and vice versa, because a centroid in $P$ can represent part of a cluster in $GT$ that has been split into two clusters in $P$, but it can also represent a cluster in $P$ that has merged two clusters in $GT$. The overall score on a video sequence can be obtained by averaging the individual scores. The proposed metric aims to ``punish'' the model both when the predictions are very far from the ground truth, and when no real groups are identified or groups that do not exist are identified. Instead, lower values for MCME will indicate that the predicted centroids match the ground truth centroids, i.e.~they represent the same clusters and are very close.

In addition, to have an easily interpretable metric from a supervised learning perspective, we present here another new metric that we call \emph{Multiple Patch Precision-Recall} (MPPR). It is based on the repeated generation of a large number of patches from the predicted and ground truth centroid density maps and a \emph{local} comparison between them, allowing for a typical classification-based evaluation mechanism. Instead of considering only the coordinates of the centroids, MPPR considers a confidence region given by the size of each patch. Let $GT_i$ and $P_i$ be the ground truth and predicted centroid density map for the $i$-th frame, respectively, and let $l_x,l_y$ be the width and height of the frame. Let $H(x,y)$ be a randomly generated point in both maps based on whose coordinates a bounding box is drawn; this bounding box acts as a sliding window that goes down and to the right. Let $w,h$ be the width and height of the rectangular bounding box, respectively. Then, MPPR computes the following quantities:
\begin{itemize}
    \item True positive ($TP$): the patches in $GT_i$ and $P_i$ are both ``active'', which means that they both contain (at least) a centroid;
    \item True negative ($TN$): the patches in $GT_i$ and $P_i$ are both ``inactive'', which means that they contain no centroid; 
    \item False positive ($FP$): the patch in $GT_i$ is inactive, while the patch in $P_i$ is active;
    \item False negative ($FN$): the patch in $GT_i$ is active, while the patch in $P_i$ is inactive.
\end{itemize}
This process is repeated $n_p$ times, where $n_p$ is usually a large number to be statistically confident that the entire frame is explored. Then, \emph{precision} and \emph{recall} for the $i$-th frame can be computed as usual. The global MPPR is obtained as the average over all video sequence frames. It should be noted that we have no guarantee that the bounding box $w \times h$ is fully contained inside the frame $l_x\times l_y$. 
The probability of obtaining a full bounding box depends solely on the random coordinates of $H(x,y)$. It can be shown that this probability amounts to $\frac{(l_x-w)(l_y-h)}{l_x l_y}$. This behavior, together with the sampling of $n_p$ bounding boxes, helps us avoid the bias we would have introduced if we had used a fixed grid of patches or a small $n_p$: in fact, in our method, a slight shift of the bounding box can result in a change of prediction (e.g., from $TP$ to $FN$), and this has the benefit of reducing the impact of a too unrealistically optimistic or pessimistic classification. In our experiments, we set $n_p = 1000$ and $w = h = 150$.

Finally, the total \textit{inference time} is calculated, which includes all the processing, from estimating the centroid density maps of two consecutive frames to thresholding and calculating the displacement of the centroids.

\subsection{Results}
Tables~\ref{tab:results} and~\ref{tab:mobilecount_results} report the accuracy results obtained by varying the FCN backbone model and the threshold $\tau$ per dataset (VisDrone CC2020 and CC2021). It is worth noting that to make the MCME metric independent of the resolution of the input frame and for a fair comparison with our previous work~\cite{IJCNN2022}, the table reports the normalized values for this metric, obtained by dividing the original value by the diagonal of the frame on which it was calculated, i.e.~the maximum possible error. Tables~\ref{tab:time} and~\ref{tab:mobilecount_time}, on the other hand, report the average time taken by the method to process two consecutive frames, depending on the backbone; having fixed the resolution, there is no difference between the datasets in terms of inference time.

A first observation that can be drawn is that not a single variant of the neural architecture performs better than the others from the point of view of predictive accuracy in all cases. Sometimes, the SBSO variant achieves the highest performance; sometimes, the multi-output models surpass the single-output one. \textcolor{black}{Similarly, there is no predominant backbone between the simple FCN and MobileCount, as they show very similar results.} As for MCME, it varies between 0.167 and 0.253 and 0.181 and 0.235 for the two datasets: since 1 is the worst possible value and lower is better, these results confirm the effectiveness of the density-based clustering strategy proposed here. \textcolor{black}{No single model exceeds the normalized MCME of 0.101 obtained in our previous preliminary study on CC2020 only, but this was achieved with a sophisticated two-stage pipeline (first density estimation, then clustering) that takes about 15 seconds to run on the same hardware~\cite{IJCNN2022}.} The slightly worse results obtained with the single-stage strategy are compensated by a much lower inference time, which is approximately 88 times shorter than in the previous work. The multi-output variants are, as expected, slightly slower than the single-output variant; however, they all showed \textit{near} real-time performance. \textcolor{black}{Notably, although MobileCount has many more parameters than the proposed simple FCN, the overall processing of the method is relatively stable, regardless of the backbone used. This suggests that the inference time of the neural network contributes only marginally to the overall time and confirms the better efficiency obtained by using a single-stage learning strategy to produce crowd density maps.}

Although MCME remains fairly stable across the different thresholds, this does not apply to precision and recall. As expected, there is a trade-off between the two metrics, with precision increasing and recall decreasing while $\tau$ increases. It is worth noting that the use of different background thresholds to filter the density maps can be set according to the specific application. For example, if the safe landing of the drone is a significant concern (as in~\cite{castellano2020crowd}), then a lower threshold may be preferred, which excludes the risk of running into false negatives. Conversely, a higher threshold can be used in video surveillance scenarios to promote better precision. The difficulty of accurately locating people in aerial scenes, which in our case translates into the problem of simultaneously maximizing precision and recall, is well-known to the community (see for example~\cite{cao2021visdrone}). \textcolor{black}{This is reflected in the better precision generally achieved by the models on CC2021, since in this dataset the scenes were acquired at a relatively lower altitude than on CC2020.}

\begin{figure}[!ht]
    \centering
    \includegraphics[width=\textwidth]{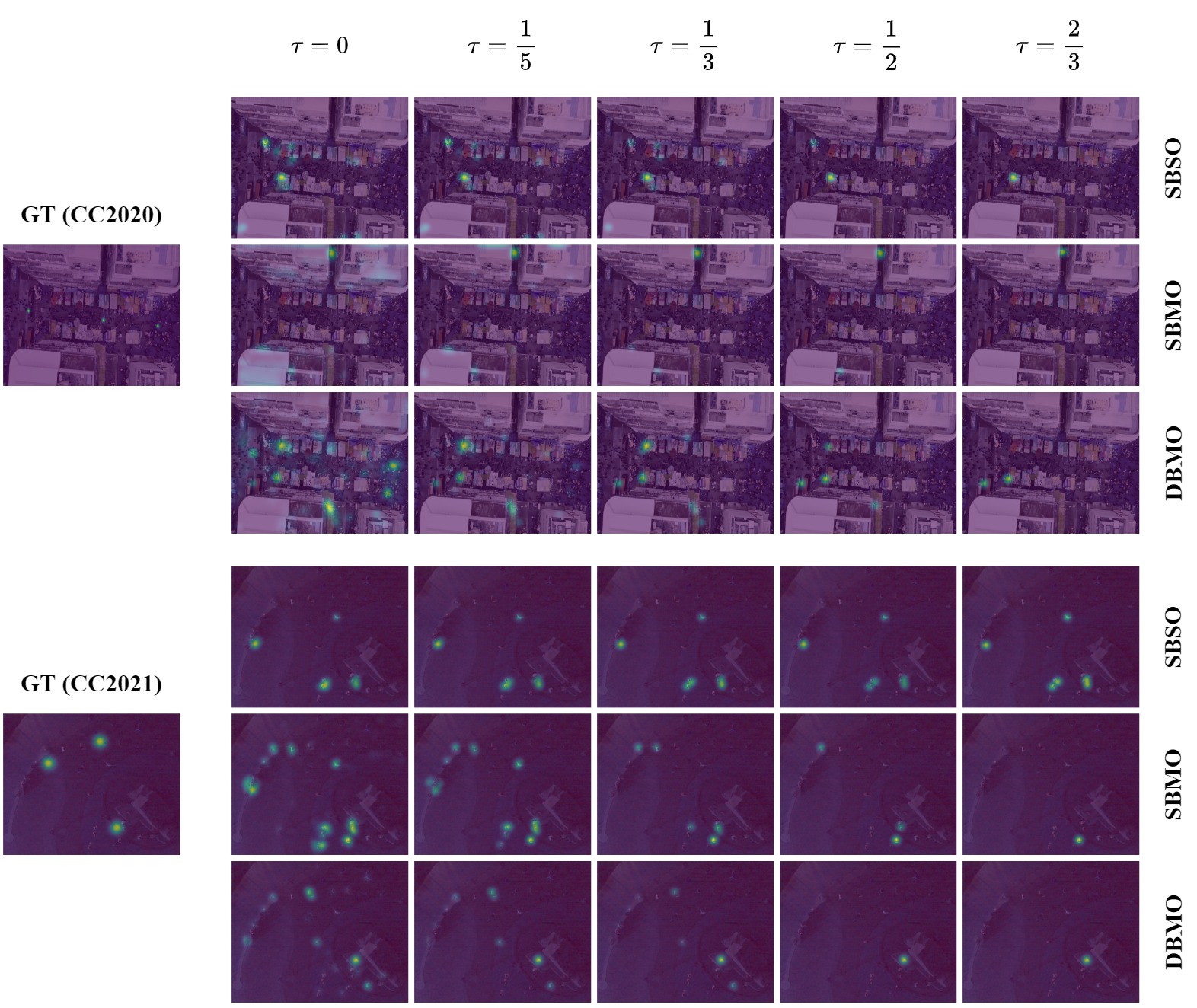}
    \caption{\textcolor{black}{Examples of centroid density maps, corresponding to ground truth maps for both CC2020 and CC2021, produced in output by the ad-hoc FCN varying the $\tau$ threshold.}}
    \label{fig:maps}
\end{figure}

\begin{figure}[!ht]
    \centering
    \includegraphics[width=\textwidth]{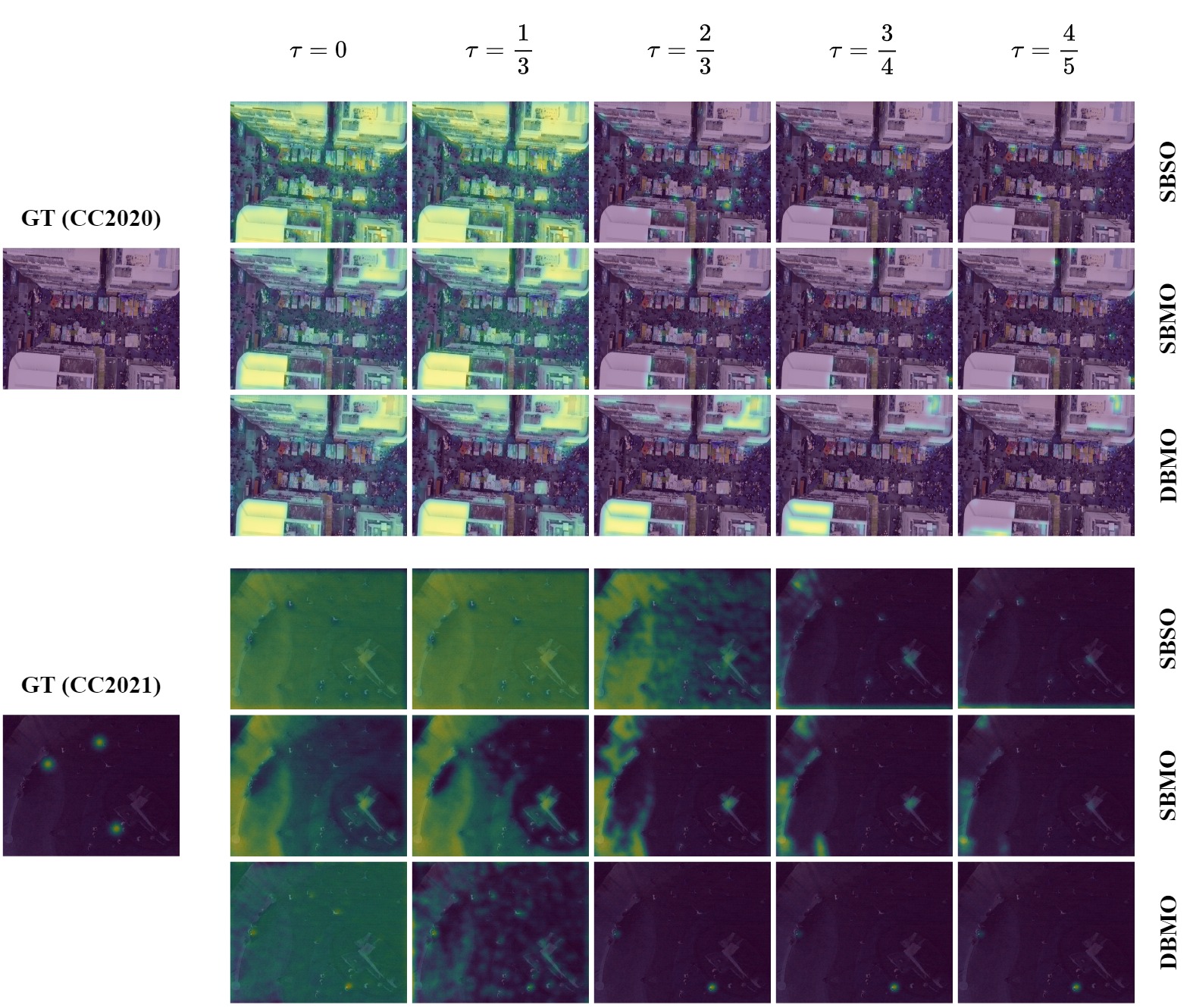}
    \caption{\textcolor{black}{Examples of centroid density maps, corresponding to ground truth maps for both CC2020 and CC2021, produced in output by MobileCount varying the $\tau$ threshold.}}
    \label{fig:maps_MC}
\end{figure}

Finally, from a qualitative point of view, we show in Figs.~\ref{fig:maps} and~\ref{fig:maps_MC} examples of centroid density maps produced in output by the ad-hoc FCN and MobileCount, respectively, by varying the $\tau$ threshold, given two ground truth maps from both datasets. The maps are superimposed on the original frames in RGB. As can be seen, they confirm the trend already observed quantitatively: the recall, i.e.~the number of centroids detected over all the centroids, tends to decrease as $\tau$ increases. \textcolor{black}{This effect is exacerbated with MobileCount, especially with zero or low $\tau$, where recall is maximum at the expense of a drastic drop in precision. Indeed, in these cases, the maps produced are pretty ineffective. This drawback could be explained by considering that MobileCount was explicitly optimized for crowd counting, thus locating each person rather than groups in the scene, which translates into a more ``conservative'' approach.} The qualitative analysis also shows how SBSO tends to adhere more to the ground truth, while SBMO and DBMO to a lesser extent. This could be explained considering that SBSO is specifically dedicated to producing centroid density maps. In contrast, in SBMO and DBMO, the crowd density estimation task can mislead the main network task. Although clusters do not perfectly match the ground truth, SBSO can still recognize groups of people within the scene, especially with higher $\tau$.

\begin{table}[H]
\centering
\scalebox{0.7}{
\begin{tabular}{ll|c|c|c|c|c|c|}
\cline{3-8}
                                                            &                             & \multicolumn{3}{c|}{\textbf{CC2020}}                                                                      & \multicolumn{3}{c|}{\textbf{CC2021}}                                                                      \\ \cline{3-8} 
                                                            &                             & \textbf{MCME}   & \textbf{Precision (\%)} & \textbf{Recall (\%)} & \textbf{MCME}   & \textbf{Precision (\%)}        & \textbf{Recall (\%)} \\ \hline
\multicolumn{1}{|l|}{\multirow{3}{*}{$\tau = 0$}}           & SBSO & \textbf{0.178}  & 26.8                    & 90.9                 & 0.198           & 51.8                           & 83.7                 \\ \cline{2-8} 
\multicolumn{1}{|l|}{}                                      & SBMO & 0.223           & 22.7                    & \textbf{97.6}        & 0.201           & 44.1                           & \textbf{100}         \\ \cline{2-8} 
\multicolumn{1}{|l|}{}                                      & DBMO & 0.203           & 22.51                   & \textbf{97.6}        & 0.201           & 44.5                           & 99.8                 \\ \hline
\multicolumn{1}{|l|}{\multirow{3}{*}{$\tau = \frac{1}{5}$}} & SBSO & 0.185           & 30                      & 76.5                 & 0.202           & 53.5                           & 78                   \\ \cline{2-8} 
\multicolumn{1}{|l|}{}                                      & SBMO & 0.229           & 22.7                    & 90.3                 & \textbf{0.181}  & 52.3                           & 86.8                 \\ \cline{2-8} 
\multicolumn{1}{|l|}{}                                      & DBMO & 0.192           & 27.2                    & 78.2                 & \textbf{0.181}  & 52.8                           & 87.2                 \\ \hline
\multicolumn{1}{|l|}{\multirow{3}{*}{$\tau = \frac{1}{3}$}} & SBSO & 0.187           & 32                      & 68.1                 & 0.208           & 54.1                           & 70.2                 \\ \cline{2-8} 
\multicolumn{1}{|l|}{}                                      & SBMO & 0.233           & 21.7                    & 76.6                 & \textbf{0.181}  & 55.3                           & 82.5                 \\ \cline{2-8} 
\multicolumn{1}{|l|}{}                                      & DBMO & 0.195           & 29.9                    & 70.6                 & 0.185           & 55.7                           & 74.3                 \\ \hline
\multicolumn{1}{|l|}{\multirow{3}{*}{$\tau = \frac{1}{2}$}} & SBSO & 0.203           & \textbf{32.9}           & 54                   & 0.217           & 54.3                           & 55.1                 \\ \cline{2-8} 
\multicolumn{1}{|l|}{}                                      & SBMO & 0.236           & 20.7                    & 52.6                 & 0.187           & 59.8                           & 70                   \\ \cline{2-8} 
\multicolumn{1}{|l|}{}                                      & DBMO & 0.208           & 32.0                    & 56.1                 & 0.201           & 58.5                           & 56.8                 \\ \hline
\multicolumn{1}{|l|}{\multirow{3}{*}{$\tau = \frac{2}{3}$}} & SBSO & 0.230           & 34.5                    & 41.5                 & 0.235           & 54.3                           & 38.1                 \\ \cline{2-8} 
\multicolumn{1}{|l|}{}                                      & SBMO & 0.253           & 19.1                    & 31.9                 & 0.202           & \textbf{63.9}                  & 51.5                 \\ \cline{2-8} 
\multicolumn{1}{|l|}{}                                      & DBMO & 0.236           & 31.9                    & 38.5                 & 0.227           & 58.7                           & 40.2                 \\ \hline
\end{tabular}}
\caption{Ad-hoc FCN effectiveness results. The best results for each individual metric per dataset are shown in bold.}\label{tab:results}

\smallskip
\smallskip

\scalebox{0.7}{
\begin{tabular}{l|c|}
\cline{2-2}
                                             & \textbf{Inference time [s]} \\ \hline
\multicolumn{1}{|l|}{SBSO}                   & \textbf{0.16}                    \\ \hline
\multicolumn{1}{|l|}{SBMO}                   & 0.17                    \\ \hline
\multicolumn{1}{|l|}{DBMO}                   & 0.18                    \\ \hline
\end{tabular}}
\caption{Ad-hoc FCN efficiency results.}\label{tab:time}
\end{table}

\begin{table}[H]
\centering
\scalebox{0.7}{
\begin{tabular}{ll|c|c|c|c|c|c|}
\cline{3-8}
                                                            &                             & \multicolumn{3}{c|}{\textbf{CC2020}}                                                                      & \multicolumn{3}{c|}{\textbf{CC2021}}                                                                      \\ \cline{3-8} 
                                                            &                             & \textbf{MCME}   & \textbf{Precision (\%)} & \textbf{Recall (\%)} & \textbf{MCME}   & \textbf{Precision (\%)}        & \textbf{Recall (\%)} \\ \hline
\multicolumn{1}{|l|}{\multirow{3}{*}{$\tau = 0$}}           & SBSO & 0.199  & 22.5                    & \textbf{100}                 & 0.205           & 44.2                           & \textbf{100}                 \\ \cline{2-8} 
\multicolumn{1}{|l|}{}                                      & SBMO & 0.201           & 22.7                    & \textbf{100}        & 0.205          & 44.1                           & \textbf{100}         \\ \cline{2-8} 
\multicolumn{1}{|l|}{}                                      & DBMO & 0.206           & 22.6                   & \textbf{100}        & 0.204           & 44.3                           & \textbf{100}                 \\ \hline
\multicolumn{1}{|l|}{\multirow{3}{*}{$\tau = \frac{1}{3}$}} & SBSO & 0.198           & 22.4                      & \textbf{100}                 & 0.205           & 43.7                           & \textbf{100}                 \\ \cline{2-8} 
\multicolumn{1}{|l|}{}                                      & SBMO & 0.201           & 22.6                    & \textbf{100}                 & \textbf{0.198}  & 44.3                           & 99.1                 \\ \cline{2-8} 
\multicolumn{1}{|l|}{}                                      & DBMO & 0.209           & 22.6                    & \textbf{100}                 & 0.202           & 44.1                           & 99.6                 \\ \hline
\multicolumn{1}{|l|}{\multirow{3}{*}{$\tau = \frac{2}{3}$}} & SBSO & 0.188           & 25.7                    & 95.2                 & 0.203           & 45.3                           & 98.1                 \\ \cline{2-8} 
\multicolumn{1}{|l|}{}                                      & SBMO & 0.184           & 24.1                    & 98.6                 & 0.204           & 49.3                  & 77.3                 \\ \cline{2-8} 
\multicolumn{1}{|l|}{}                                      & DBMO & 0.231           & 23.1                    & 98.0                 & 0.186           & 59.1                           & 64.8                 \\ \hline
\multicolumn{1}{|l|}{\multirow{3}{*}{$\tau = \frac{3}{4}$}} & SBSO & 0.213           & \textbf{40.9}                    & 45.5                 & 0.194           & 46.7                           & 93.9                 \\ \cline{2-8} 
\multicolumn{1}{|l|}{}                                      & SBMO & 0.170           & 27.3                    & 93.1                 & 0.215           & 47.9                  & 62.5                 \\ \cline{2-8} 
\multicolumn{1}{|l|}{}                                      & DBMO & 0.230           & 23.3                    & 87.5                 & 0.20           & 61.5                           & 49.1                 \\ \hline
\multicolumn{1}{|l|}{\multirow{3}{*}{$\tau = \frac{4}{5}$}} & SBSO & 0.176           & 30.6                    & 88.3                 & 0.193           & 48.1                           & 85.7                 \\ \cline{2-8} 
\multicolumn{1}{|l|}{}                                      & SBMO & \textbf{0.167}           & 30.2                    & 86.2                 & 0.224           & 46.4                  & 52.0                 \\ \cline{2-8} 
\multicolumn{1}{|l|}{}                                      & DBMO & 0.225           & 23.7                    & 79.0                 & 0.211           & \textbf{64.4}                           & 41.9                 \\ \hline
\end{tabular}}
\caption{\textcolor{black}{MobileCount effectiveness results. The best results for each individual metric per dataset are shown in bold.}}\label{tab:mobilecount_results}

\smallskip
\smallskip

\scalebox{0.7}{
\begin{tabular}{l|c|}
\cline{2-2}
                                             & \textbf{Inference time [s]} \\ \hline
\multicolumn{1}{|l|}{SBSO}                   & \textbf{0.15}                    \\ \hline
\multicolumn{1}{|l|}{SBMO}                   & 0.18                    \\ \hline
\multicolumn{1}{|l|}{DBMO}                   & 0.16                    \\ \hline
\end{tabular}}
\caption{\textcolor{black}{MobileCount efficiency results.}}\label{tab:mobilecount_time}
\end{table}


\section{Conclusion}
\label{sec:conclusion}
In this article, we tackled the problem of crowd flow detection from drones, which was still an unexplored research direction. Such a system can be helpful for various security and management applications, especially in smart city scenarios. In particular, the joint exploitation of crowd density estimation and clustering within a video sequence shot by a drone provided encouraging results, especially from the point of view of efficiency, which can be crucial in critical tasks.

Future development of the research presented in this paper could be to increase the size of the dataset by integrating the available scenes with synthetic data, as done, for example, in~\cite{zhao2020flow}. Such a strategy can help further improve the robustness of the deep learning model. 
\textcolor{black}{Second, it is worth noting that the proposed framework assumes the availability of video sequences shot by a drone, but the neural network is fed with one frame at a time. A different strategy to be explored in the future could be to feed the network with the video sequence to account for the temporal information directly in the model.}
\textcolor{black}{Third, provided adequate ground truth is available, the method could also be used for other similar tasks, such as vehicle counting/tracking.}
Finally, another future work concerns the experimentation of the method on new real-world situations aboard a drone to best calibrate the parameters considered. Testing the generalizability of the method to different urban and non-urban contexts can increase confidence in UAV technology.

\paragraph*{Acknowledgment}
This work was supported by the Italian Ministry of University and Research within the ``RPASInAir'' project under grant PON ARS01\_00820.

\bibliographystyle{elsarticle-num} 
\bibliography{cas-refs}

\begin{thebibliography}{10}
\expandafter\ifx\csname url\endcsname\relax
  \def\url#1{\texttt{#1}}\fi
\expandafter\ifx\csname urlprefix\endcsname\relax\def\urlprefix{URL }\fi
\expandafter\ifx\csname href\endcsname\relax
  \def\href#1#2{#2} \def\path#1{#1}\fi

\bibitem{sindagi2018survey}
V.~A. Sindagi, V.~M. Patel, A survey of recent advances in {CNN}-based single
  image crowd counting and density estimation, Pattern Recognition Letters 107
  (2018) 3--16.

\bibitem{li2021approaches}
B.~Li, H.~Huang, A.~Zhang, P.~Liu, C.~Liu, Approaches on crowd counting and
  density estimation: a review, Pattern Analysis and Applications (2021) 1--22.

\bibitem{akbari2021applications}
Y.~Akbari, N.~Almaadeed, S.~Al-maadeed, O.~Elharrouss, Applications, databases
  and open computer vision research from drone videos and images: a survey,
  Artificial Intelligence Review 54~(5) (2021) 3887--3938.

\bibitem{tzelepi2019graph}
M.~Tzelepi, A.~Tefas, Graph embedded convolutional neural networks in human
  crowd detection for drone flight safety, IEEE Transactions on Emerging Topics
  in Computational Intelligence 5~(2) (2019) 191--204.

\bibitem{zhu2021detection}
P.~Zhu, L.~Wen, D.~Du, X.~Bian, H.~Fan, Q.~Hu, H.~Ling, Detection and tracking
  meet drones challenge (2021).
\newblock \href {http://arxiv.org/abs/2001.06303} {\path{arXiv:2001.06303}}.

\bibitem{kok2016crowd}
V.~J. Kok, M.~K. Lim, C.~S. Chan, Crowd behavior analysis: A review where
  physics meets biology, Neurocomputing 177 (2016) 342--362.

\bibitem{IJCNN2022}
G.~Castellano, C.~Mencar, G.~Sette, F.~S. Troccoli, G.~Vessio, Crowd flow
  detection from drones with fully convolutional networks and clustering, in:
  2022 International Joint Conference on Neural Networks (IJCNN 2022), IEEE,
  2022, pp. 1--8.

\bibitem{du2020visdrone}
D.~Du, L.~Wen, P.~Zhu, H.~Fan, Q.~Hu, H.~Ling, M.~Shah, J.~Pan, A.~Al-Ali,
  A.~Mohamed, et~al., Visdrone-cc2020: The vision meets drone crowd counting
  challenge results, in: European Conference on Computer Vision, Springer,
  2020, pp. 675--691.

\bibitem{liu2021visdrone}
Z.~Liu, Z.~He, L.~Wang, W.~Wang, Y.~Yuan, D.~Zhang, J.~Zhang, P.~Zhu,
  L.~Van~Gool, J.~Han, et~al., Vis{D}rone-{CC}2021: The vision meets drone
  crowd counting challenge results, in: Proceedings of the IEEE/CVF
  International Conference on Computer Vision, 2021, pp. 2830--2838.

\bibitem{lan2018pedestrian}
W.~Lan, J.~Dang, Y.~Wang, S.~Wang, Pedestrian detection based on yolo network
  model, in: 2018 IEEE international conference on mechatronics and automation
  (ICMA), IEEE, 2018, pp. 1547--1551.

\bibitem{molchanov2017pedestrian}
V.~Molchanov, B.~Vishnyakov, Y.~Vizilter, O.~Vishnyakova, V.~Knyaz, Pedestrian
  detection in video surveillance using fully convolutional {YOLO} neural
  network, in: Automated visual inspection and machine vision II, Vol. 10334,
  International Society for Optics and Photonics, 2017, p. 103340Q.

\bibitem{gao2020cnn}
G.~Gao, J.~Gao, Q.~Liu, Q.~Wang, Y.~Wang, {CNN}-based density estimation and
  crowd counting: A survey, arXiv preprint arXiv:2003.12783 (2020).

\bibitem{zhu2020crowd}
L.~Zhu, C.~Li, Z.~Yang, K.~Yuan, S.~Wang, Crowd density estimation based on
  classification activation map and patch density level, Neural Computing and
  Applications 32~(9) (2020) 5105--5116.

\bibitem{zhang2020cross}
G.~Zhang, Y.~Pan, L.~Zhang, R.~L.~K. Tiong, Cross-scale generative adversarial
  network for crowd density estimation from images, Engineering Applications of
  Artificial Intelligence 94 (2020) 103777.

\bibitem{sindagi2017cnn}
V.~A. Sindagi, V.~M. Patel, {CNN}-based cascaded multi-task learning of
  high-level prior and density estimation for crowd counting, in: 2017 14th
  IEEE International Conference on Advanced Video and Signal Based Surveillance
  (AVSS), IEEE, 2017, pp. 1--6.

\bibitem{jiang2021shufflecount}
M.~Jiang, J.~Lin, Z.~J. Wang, Shuffle{C}ount: Task-specific knowledge
  distillation for crowd counting, in: 2021 IEEE International Conference on
  Image Processing (ICIP), IEEE, 2021, pp. 999--1003.

\bibitem{jiang2021smartly}
M.~Jiang, J.~Lin, Z.~J. Wang, A smartly simple way for joint crowd counting and
  localization, Neurocomputing 459 (2021) 35--43.

\bibitem{liang2021transcrowd}
D.~Liang, X.~Chen, W.~Xu, Y.~Zhou, X.~Bai, Trans{C}rowd: Weakly-supervised
  crowd counting with transformer, arXiv preprint arXiv:2104.09116 (2021).

\bibitem{dosovitskiy2020image}
A.~Dosovitskiy, L.~Beyer, A.~Kolesnikov, D.~Weissenborn, X.~Zhai,
  T.~Unterthiner, M.~Dehghani, M.~Minderer, G.~Heigold, S.~Gelly, et~al., An
  image is worth 16x16 words: Transformers for image recognition at scale,
  arXiv preprint arXiv:2010.11929 (2020).

\bibitem{castellano2020crowd}
G.~Castellano, C.~Castiello, C.~Mencar, G.~Vessio, Crowd detection in aerial
  images using spatial graphs and fully-convolutional neural networks, IEEE
  Access 8 (2020) 64534--64544.

\bibitem{gajjar2017human}
V.~Gajjar, A.~Gurnani, Y.~Khandhediya, Human detection and tracking for video
  surveillance: A cognitive science approach, in: Proceedings of the IEEE
  international conference on computer vision workshops, 2017, pp. 2805--2809.

\bibitem{xiao2019human}
Y.~Xiao, V.~R. Kamat, C.~C. Menassa, Human tracking from single {RGB-D} camera
  using online learning, Image and Vision Computing 88 (2019) 67--75.

\bibitem{yan2020online}
Z.~Yan, T.~Duckett, N.~Bellotto, Online learning for 3{D} {L}i{DAR}-based human
  detection: Experimental analysis of point cloud clustering and classification
  methods, Autonomous Robots 44~(2) (2020) 147--164.

\bibitem{wen2021detection}
L.~Wen, D.~Du, P.~Zhu, Q.~Hu, Q.~Wang, L.~Bo, S.~Lyu, Detection, tracking, and
  counting meets drones in crowds: A benchmark, in: Proceedings of the IEEE/CVF
  Conference on Computer Vision and Pattern Recognition, 2021, pp. 7812--7821.

\bibitem{chebil2022toward}
K.~Chebil, S.~Htiouech, M.~Khemakhem, Toward optimal periodic crowd tracking
  via unmanned aerial vehicles, Computers \& Industrial Engineering (2022).

\bibitem{comaniciu2002mean}
D.~Comaniciu, P.~Meer, Mean shift: A robust approach toward feature space
  analysis, IEEE Transactions on pattern analysis and machine intelligence
  24~(5) (2002) 603--619.

\bibitem{zhang2016single}
Y.~Zhang, D.~Zhou, S.~Chen, S.~Gao, Y.~Ma, Single-image crowd counting via
  multi-column convolutional neural network, in: Proceedings of the IEEE
  conference on computer vision and pattern recognition, 2016, pp. 589--597.

\bibitem{long2015fully}
J.~Long, E.~Shelhamer, T.~Darrell, Fully convolutional networks for semantic
  segmentation, in: Proceedings of the IEEE conference on computer vision and
  pattern recognition, 2015, pp. 3431--3440.

\bibitem{wang2020mobilecount}
P.~Wang, C.~Gao, Y.~Wang, H.~Li, Y.~Gao, Mobile{C}ount: An efficient
  encoder-decoder framework for real-time crowd counting, Neurocomputing 407
  (2020) 292--299.

\bibitem{sandler2018mobilenetv2}
M.~Sandler, A.~Howard, M.~Zhu, A.~Zhmoginov, L.-C. Chen, Mobile{N}et{V}2:
  Inverted residuals and linear bottlenecks, in: Proceedings of the IEEE
  conference on computer vision and pattern recognition, 2018, pp. 4510--4520.

\bibitem{nekrasov2018light}
V.~Nekrasov, C.~Shen, I.~Reid, Light-{W}eight {R}efine{N}et for real-time
  semantic segmentation, arXiv preprint arXiv:1810.03272 (2018).

\bibitem{cao2021visdrone}
Y.~Cao, Z.~He, L.~Wang, W.~Wang, Y.~Yuan, D.~Zhang, J.~Zhang, P.~Zhu,
  L.~Van~Gool, J.~Han, et~al., Vis{D}rone-{DET}2021: The vision meets drone
  object detection challenge results, in: Proceedings of the IEEE/CVF
  International Conference on Computer Vision, 2021, pp. 2847--2854.

\bibitem{zhao2020flow}
Z.~Zhao, T.~Han, J.~Gao, Q.~Wang, X.~Li, A flow base bi-path network for
  cross-scene video crowd understanding in aerial view, in: European Conference
  on Computer Vision, Springer, 2020, pp. 574--587.

\end{thebibliography}





\end{document}